\pdfoutput=1
\PassOptionsToPackage{backref=page}{hyperref}
\documentclass[11pt]{article}

\newif\ifauthordecided
\authordecidedtrue %

\newif\ifarxiv
\arxivtrue %

\newif\ifperfect
\perfectfalse

\ifarxiv
    \usepackage{acl}
\else
    \usepackage[review]{acl}
\fi

\usepackage{times}
\usepackage{latexsym}
\usepackage{pgfplots}

\pgfplotsset{compat=1.17}

\usepackage[T1]{fontenc}

\usepackage[utf8]{inputenc}
\usepackage{makecell}
\usepackage{booktabs} %
\usepackage{multirow} %

\usepackage{microtype}

\usepackage{sec/zhijing_package} %

\newcommand{\basemodel}{\textsc{Base}\xspace}
\newcommand{\ourmodel}{{\textsc{AmrCoT}}\xspace}

\title{
Analyzing the Role of
Semantic Representations \\
in the Era of Large Language Models
}

\ifauthordecided
\author{Zhijing Jin\thanks{\hspace{0.1cm} Equal contribution.}
\\
  MPI \& ETH \\
  {\small\texttt{jinzhi@ethz.ch}} \\\And
  Yuen Chen\samethanks \\
  UIUC \\
  {\small\texttt{yuenc2@illinois.edu}} \\\And
  Fernando Gonzalez\samethanks \\
  ETH \\
  {\small\texttt{fer.adauto@gmail.com}} \\\And
  Jiarui Liu \\
  CMU \\
  {\small\texttt{jiarui@cmu.edu}} 
\\\AND
  Jiayi Zhang \\
  University of Michigan \\
  {\small\texttt{jiayizzz@umich.edu}} \\\And
  Julian Michael \\
  NYU \\
  {\small\texttt{julianjm@nyu.edu}} \\\And
  Bernhard Sch\"olkopf \\
  MPI \\
  {\small\texttt{bs@tue.mpg.de}}
  \\\And
  Mona Diab \\
  CMU \\
  {\small\texttt{mdiab@cs.cmu.edu}} \\
}
\fi

\begin{document}

\maketitle
\begin{abstract}
Traditionally, natural language processing (NLP) models often use a rich set of features created by linguistic expertise, such as semantic representations. 
However, in the era of large language models (LLMs), more and more tasks are turned into generic, end-to-end sequence generation problems. In this paper, we investigate the question: what is the role of
semantic representations in the era of LLMs?
Specifically, we investigate the effect of Abstract Meaning Representation (AMR) across five diverse NLP tasks. We propose an AMR-driven chain-of-thought prompting method, which we call \ourmodel, and find that it generally hurts performance more than it helps.
To investigate what AMR may have to offer on these tasks, we conduct a series of analysis experiments. We find that it is difficult to predict which input examples AMR may help or hurt on, but errors tend to arise with multi-word expressions, named entities, and in the final inference step where the LLM must connect its reasoning over the AMR to its prediction. We recommend focusing on these areas for future work in semantic representations for LLMs.%
\footnote{\ifarxiv
Our code: {\href{https://github.com/causalNLP/amr_llm}{https://github.com/causalNLP/amr\_llm}}.
\fi
}

\end{abstract}

\section{Introduction}

\begin{figure}[ht]
    \centering
    \includegraphics[width=\linewidth]{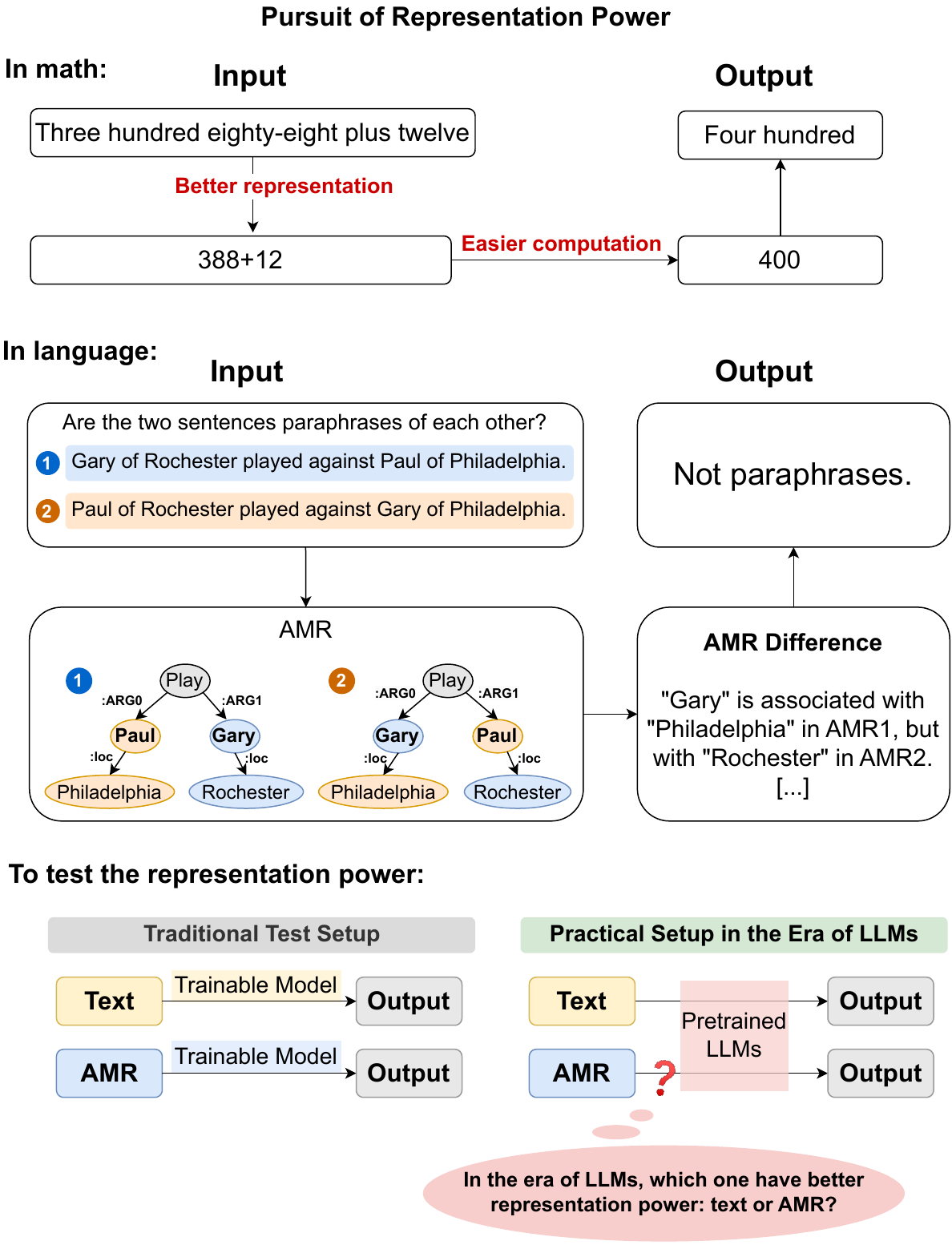}
    \caption{
    The role of representation power in different fields. Analogous
    to Arabic numbers for math, AMR is designed to efficiently and explicitly represent the semantic features of text. Existing work using AMR is concerned with trainable models,
    whereas we investigate the use of AMR in the modern practical setup of pre-trained LLMs.
    }
    \label{fig:intro}
\end{figure}

Formal representations of linguistic structure and meaning have long held an important role in the construction and evaluation of NLP systems. Semantic representations such as Abstract Meaning Representation \citep[AMR;][]{banarescu2013abstract} are designed to distill the semantic information of text to a graph consisting of the entities, events, and states mentioned in a text and the relations between them.
Existing studies have shown the benefits of representations like AMR in a variety of NLP tasks, such as 
paraphrase detection \citep{Issa2018Abstract},
machine translation \citep{song2019semantic}, event extraction \citep{Garg2015Extracting,huang2018zero}, code generation \citep{yin2017syntactic}, 
and others \citep{Dohare2017Text,jangra2022star,wolfson2020break,kapanipathi2021leveraging}.
By explicitly representing the propositional structure of sentences, AMR removes much of the information from text that is irrelevant to semantic tasks, while surfacing the most important information (entities, relations, etc.), rendering them easier to operate on. In theory, this implies that using AMR as an intermediate representation should make it easier for a model to learn to perform such tasks, in the same way that a representation like Arabic numerals aids with arithmetic (see \cref{fig:intro}).

However, learning to produce and operate over representations like AMR is nontrivial, especially since AMR data is limited.
In contrast, modern NLP systems based on large language models (LLMs) learn to directly manipulate text very effectively \cite{ignat-etal-2024-has},
not only achieving high performance on a variety of tasks without using intermediate formal representations~\citep{brown2020gpt3},
but also achieving gains by directly using informal \textit{textual} intermediates in methods such as chain-of-thought (CoT) prompting~\citep{wei2022chain}.
Due to economic concerns \citep{zhao2022greener,samsi2023words,patterson2021carbon,sharir2020cost},
there is a growing trend to utilize readily available pre-trained LLMs in various application scenarios, without allocating additional resources for training or fine-tuning the models.
These trends raise the question:
\begin{quote}
    \textit{What is the role of semantic representations in the era of LLMs, when no training or finetuning is involved?}
\end{quote} 

Motivated by this question, we propose a theoretical formulation of representation power, and what it means to have an ideal representation for text, using ideas from Kolmogorov complexity \cite{solomonoff1964formal,kolmogorov1965three}. Our key observation is that making use of even a very strong intermediate representation requires optimizing the model with regard to that representation; however, when using (out of the box) pretrained LLMs, the optimal representation will be the one which the LLM can \textit{most effectively use} on the basis of its pretraining, which might shift away from the optimal representation for a learnable mapping to the output space.
In short, the \textit{a priori} ideal representation for a task is not necessarily the ideal representation for an LLM to use.

Given this, we empirically study how good AMR is as an intermediate representation for LLMs.
Specifically, we answer the following three questions:
(1) Does AMR help LLM performance?
(2) When does AMR help, and when doesn't it?
(3) Why does it help or not help in these cases?

On a diverse set of {five} NLP tasks, our experiments show that the contribution of AMR in the era of LLMs is not as great as that in the traditional setup where we can optimize the model for the representation. AMR causes a slight fluctuation of performance by -3 to +1 percentage points. However, we find that AMR is helpful for a subset of samples.
We also find that the next step for using AMR to improve performance is likely not improving AMR parser performance, but
improving the LLM's ability to map AMR representations into the output space.

In summary, the contributions of our work are as follows:
\begin{enumerate}
    \item 
    We are the first to investigate how semantic representations such as AMR can be leveraged to help LLM performance in the practical situation where no training is involved;
    \item 
    We propose a formalization of representation power for intermediate representations of language, and comprehensive experimental studies investigating \textit{whether}, \textit{when}, and \textit{why} AMR can help when performing semantic tasks with LLMs;
    \item We present thorough analysis experiments and results reflecting on the contribution of traditional linguistic structures such as semantic representations in the current wave of LLMs, and point out potential areas for improvement.
\end{enumerate}

\section{Formalizing Representation Power}
\label{sec:repr_power}

In this section, we propose a framework to formulate representation power in both the pre-LLM era, where we do not outsource the training of the models, and the LLM era, where a lot of practical settings are to optimize the representation with regard to given fixed LLMs.
\subsection{Notation}
Suppose we have a dataset $D:= \{(x_i, y_i)\}_{i=1}^N$ consisting of $N$ pairs of input $x_i$ and corresponding output $y_i$. 
Given the task to learn
the $x \mapsto y$ mapping, we can consider it as a two-stage modeling process: the first step is to convert the raw input $x$ into a good representation $r$ by the representation model $g: x \mapsto r
$,
and the second step is to perform the computation that takes the representation $r$ and predicts the output  $y$ by a computation model $h: r \mapsto y$. In this way, we decompose the resulting
overall $x \mapsto y$ modeling process into
\begin{align}
p(y|x) & = p_g(r|x) p_h(y|r)
~,\label{eq:chain}
\end{align} where the overall probabilistic model $p(y|x)$ is turned into first the representation step $p_g(r|x)$ to generate the representation $r$ given the input $x$, and then the computation step $p_h(y|r)$ to perform operations on the intermediate representation $r$ to derive the output $y$.

\subsection{Problem Formulation}
Let us draw some intuition from the math example in \cref{fig:intro}. To represent numbers, the first representation choice $r_1$ is the English expression, such as ``Three hundred eighty-eight plus twelve,'' and the second, intuitively stronger representation $r_2$ is the same calculation in Arabic numbers, ``388+12''. For our research question of whether AMR demonstrates stronger representation power than
raw text, we formulate the question as follows:
\begin{itemize}
    \item Representation choices: $\mathcal{R} = \{ \mathrm{text}, \mathrm{amr} \}$ (which is an instance of text and its semantic representation, a common question of interest in linguistics);
    \item Representation power: some properties of the function $h: r \mapsto y$.
\end{itemize}
The next question becomes theorizing \textit{what} properties of the computation model $h: r \mapsto y$ we are optimizing for, for which we will introduce two formulations, one in the pre-LLM era and the other in the LLM era.

\subsection{Ideal Representations in the Pre-LLM Era}
\label{sec:ideal}

Continuing on with the Arabic number example:
Our claim is that the representation $r_2$ (i.e., the Arabic number) is better than $r_1$ (i.e., the English expression) because the computation for $h_2: r_2 \mapsto y$ is simpler than $h_1: r_1 \mapsto y$, as measured by \textit{Kolmogorov complexity}, or \textit{algorithmic entropy} \cite{solomonoff1964formal,kolmogorov1965three}, which is a theoretical construct of the complexity of an algorithm in bits.\footnote{Formally, Kolmogorov complexity is the length of the shortest program which produces a string.} Intuitively, the shortest program specifying an algorithm to take English expressions like ``Three hundred eighty-eight plus twelve'' as input and produce ``Four hundred'' as output should be longer than for the one taking ``388+12'' as input and ``400'' as output, since the former requires more complicated string manipulation to achieve the same effect.

We also use this notion of Kolmogorov complexity to quantify the power of representations for language. The intuition is that powerful representations are those that significantly simplify the complexity of the computation model $h$.
Hence, the optimal representation function $g^* \in \mathcal{G}$ from the set $\mathcal{G}$ of possible functions should satisfy
\begin{align}
    g^* = \argmin_{g\in \mathcal{G}} \min_{h\in \mathcal{H}} K(h)
    ~,\label{eq:r}
    \\
    \hfill \text{where } h: g(x) \mapsto y
    ~,\label{eq:perfect_compute}    
\end{align}
and the optimal representation $r^*$ is
\begin{align}
    r^* = g^*(x)
    ~.
    \label{eq:r_res}
\end{align}
Here, note that given each representation function, we optimize over all possible computation models $h$ from the hypothesis space $\mathcal{H}$ to achieve the minimal Kolmogorov complexity $K(h)$.

If the representation enables the computation model $h$ to have
a low Kolmogorov complexity, it usually results in several good properties,
such as that learning $h$ has smaller generalization risks, requires fewer data samples, %
has smaller empirical risks, and 
results in more robustness, as introduced in \citet{jin-etal-2021-causal}. 
Various studies explore the theoretical foundations for the above claims by connecting Kolmogorov complexity with statistical learning theory. For example, \citet[§4.6.1]{vapnik1995nature} shows that an upper bound of Kolmogorov complexity, called ``compression coefficient,'' can bound the generalization error in classification problems; 
\citet[Eq.~3]{shalevshwartz2014understanding} and \citet{goldblum2023free} show that generalization error is upper bounded by training error plus a term involving the Kolmogorov complexity of the hypothesis space.

This is, we argue, the implicit framework behind many previous studies showing AMR as a better representation than the raw text sequence by demonstrating its better performance \citep{turian2010word}, data efficiency \citep{liu2102representation}, and robustness and domain transferability \citep{li2016learning,jin-etal-2021-causal}.
A crucial element of these studies is that they train models customized explicitly for the AMR representation, optimizing $h$ over the hypothesis space $\mathcal{H}$.

\subsection{Representation Power in the LLM Era}\label{sec:llm_repr}

As mentioned previously, in the era of LLMs, we are moving towards the paradigm where the model training is usually outsourced, and during the inference stage, i.e., for most use cases, the model weights are fixed.
Formally, this means two differences from the previous setting: (1) the hypothesis space $\mathcal{H}$ is collapsed to a size of one, containing only the fixed function $h_{\mathrm{LLM}}$, (2) the optimization constraint in \cref{eq:perfect_compute} that $h$ can map the representation to the ground truth $y$ is not necessarily guaranteed, namely that $h$ could lead to $\hat{y}$, with certain estimation error.

Therefore, the key measure of representation power in the LLM era naturally shifts from simplicity of the computation model $h$---which aids optimization towards low estimation error---to low estimation error itself, i.e., $\mathbb{E}[\delta(\hat{y}, y)]$, where $\delta$ is the error function.
This change results in a shift from the double optimization over both $r$ and $h$ to the optimization only of $r$ with regard to the fixed $h_{\mathrm{LLM}}$:
\begin{align}
    g^*_{\mathrm{LLM}} &= \argmin_{g\in \mathcal{G}} \mathbb{E}[\delta(\hat{y} , y)]  
    ~,\label{eq:r_llm_error}
    \\
    &= \argmin_{g\in \mathcal{G}} \mathbb{E}[\delta(h_{\mathrm{LLM}}(g(x)) , y)]
    ~,\label{eq:r_llm}
\end{align}
where the optimal representation $r^*_{\mathrm{LLM}}$ becomes
\begin{align}
    r^*_{\mathrm{LLM}} = g^*_{\mathrm{LLM}}(x)
    ~.\label{eq:r_llm_res}
\end{align} 
This framework can also be used to explain the success, for example, of CoT prompting \cite{wei2022chain,nye2021show} in terms of how the intermediate representation generated by CoT better unlocks the power of LLMs.

Comparing \cref{eq:r,eq:perfect_compute,eq:r_res} with \cref{eq:r_llm,eq:r_llm_error,eq:r_llm_res}, we can see that the ideal best representation $r^*$ is not necessarily equal to the representation $r^*_{\mathrm{LLM}}$ that works well with LLMs, so there remains a need for experiments to fill in this knowledge gap.

It is also worth noting that for any learned representation function $g$, errors in $p_g(r|x)$ relative to $p_{g^*}(r|x)$ may cascade into the computation step $p(y|r)$, harming the final output. We investigate this concern in \cref{sec:gold}.

\section{Designing the \ourmodel Experiments}\label{sec:exp}

We introduce an AMR-driven prompting method which we call \ourmodel, and investigate its performance on five datasets with five LLMs.
\subsection{Dataset Setup}

\begin{table}[ht]
    \centering
    \small
    \setlength\tabcolsep{6pt}
    \begin{tabular}{llccccccc}
    \toprule
    Dataset & Task
    & Test Size
    \\ \midrule
    PAWS & Paraphrase Detection  & 8,000 \\
    WMT16 & Translation & 5,999 \\
    Logic & Logical Fallacy Detection   & 2,449 \\
    Pubmed45 & Event Extraction&  5,000 \\
    SPIDER & Text2SQL Code Generation & 8,034 \\
    \bottomrule
    \end{tabular}
    \caption{Tasks and datasets  used.
    }
    \label{tab:data_size}
\end{table}
We test \ourmodel\ on paraphrase detection \citep{zhang2019paws}, machine translation \citep{bojar2016wmt1}, logical fallacy detection \cite{jin-etal-2022-logical}, event extraction \citep{Garg2015Extracting}, and text-to-SQL generation \citep{yu2018spider}.
We select these tasks as they hinge on complex sentence structures and
most of them are reported to have benefited from AMR in the pre-LLM era \cite{Issa2018Abstract,song2019semantic,Garg2015Extracting,yin2017syntactic}.

For each dataset,
we first take the entire original test set, and if it has fewer than  5,000 examples, we also include the development or training set.
Data statistics are in \cref{tab:data_size} and details on test set construction are in \cref{appd:split}.

\subsection{\ourmodel Prompt Design}

To test the utility of AMR with LLMs,
we draw inspiration from the CoT prompt design \cite{wei2022chain,nye2021show}, together with CoT variants on causal \cite{jin2023cladder} and moral reasoning tasks \cite{jin2022make}, which enables models to answer an initially difficult question with the help of assistive intermediate steps to 
render the task easier. 

We propose \ourmodel, in which we supplement the input text with an automatically-generated AMR and condition the LLM on the input text and AMR when generating the answer.
If AMR has a stronger representation power than the raw text, then providing AMR as an assistive intermediate step should improve the performance of LLMs.
\begin{table}[b]
    \centering
\small
    \setlength\tabcolsep{2pt}
    \begin{tabular}{lp{6.5cm}}
\toprule
\basemodel & Please translate the following text from English to German.\newline Text:  \texttt{\{sentence1\}}\newline Translation:
\\ \midrule
\ourmodel & You are given a text and its abstract meaning representation (AMR).\newline Text: \texttt{\{sentence1\}}\newline \textbf{AMR:}\newline \textbf{\texttt{\{amr1\}}}
\newline Please translate the text from English to German. \textit{You can refer to the provided AMR if it helps you in creating the translation.}\newline Translation:
\\ 
\bottomrule
    \end{tabular}
    \caption{Example \basemodel and \ourmodel prompt (for the translation task). We serialize AMRs with the commonly used Penman notation \cite{patten-1993-book}.}
    \label{tab:prompt_ex}
\end{table}

We compare \ourmodel to directly querying the LLMs, denoted \basemodel. An example prompt pair is shown in \cref{tab:prompt_ex}, and all prompts for all datasets are in \cref{appd:prompts}.

\subsection{Language Models}
Since our experiments require models that can reasonably understand and reason over the symbols in AMRs, we find that only the
instruction-tuned GPT models, from \textit{text-davinci-001} to GPT-4 are capable of processing it, but not the open-sourced models such as LLaMa and Alpaca, at the time we conducted our research.
For reproducibility, we set the text generation temperature to 0 for all models, and we use the model checkpoints from June 13, 2023 for GPT-3.5 and GPT-4, namely \textit{gpt-3.5-turbo-0613} and \textit{gpt-4-0613}.

\subsection{Addressing Research Questions}

\section{Q1: Does AMR Help LLMs?}\label{sec:q1}
First, we are interested in the utility of AMR as an intermediate representation for LLMs. Specifically, we answer the following subquestions: what is the overall effect of AMR as a representation on LLMs' performance (\cref{sec:overall_effect})? Does the effect vary case by case (\cref{sec:help_harm_rate})? And how does the effect change with using various LLMs with different levels of capabilities (\cref{sec:model_version})?

\subsection{Overall Effect of AMR}\label{sec:overall_effect}

We first evaluate the overall effect of AMR as a representation to assist LLMs. Following the setup in \cref{sec:exp}, \cref{tab:main_res} shows performance on our five NLP tasks. Comparing \ourmodel to the \basemodel method which directly queries LLMs, AMR does not have an overall positive impact on performance. The performance fluctuates between a slight drop (-1 to -3 in most tasks) and a slight increase (+0.61 in the case of Text-to-SQL code generation).

\begin{table}[ht]
    \centering
    \small
    \setlength\tabcolsep{6pt}
    \begin{tabular}{llccccccc}
    \toprule
    Dataset & Task
    & \basemodel & $\Delta$\ourmodel
    \\ \midrule
    PAWS & Paraphrase Detection  & 78.25 & -3.04 \\
    WMT & Translation & 27.52 & -0.83 \\
    Logic & Fallacy Detection   & 55.61 & -0.49 \\
    Pubmed45 & Event Extraction&  39.65 & -3.87\\
    SPIDER & Text2SQL & 43.78 & +0.61 \\
    \bottomrule
    \end{tabular}
    \caption{
    Across the five tasks, we report the baseline performance (\basemodel), and the additional impact of \ourmodel ($\Delta$\ourmodel), using GPT-4. See statistical significance tests in \cref{appd:stats_signi}.
    }
    \vspace{-.5em}
    \label{tab:main_res}
\end{table}

\subsection{Helpfulness of AMR in Some Cases}\label{sec:help_harm_rate}

Using AMR hardly changes overall performance, but this could be either because it does not change model predictions or because it helps in roughly as many cases as it hurts. To explore which is the case, we calculate the percentage of examples which are helped and hurt by \ourmodel, shown in \cref{tab:help_hurt_rate}. We count a sample as helped by AMR if its prediction improves (i.e., the output changes from incorrect to correct in classification tasks, or its score increases in text generation tasks), and hurt by AMR if its prediction degrades; the rest of the examples are considered unchanged.
\begin{table}[t]
    \centering
    \small
    \setlength\tabcolsep{6pt}
    \begin{tabular}{lccccccc}
    \toprule
    Dataset
    & \% Helped & \% Hurt & \% Unchanged
    \\ \midrule
    PAWS &
    16.48 & 20.16 & 63.36 \\
    WMT &
     16.45 &21.17 & 62.38 \\
    Logic &
     1.96 & 2.45 & 95.59 \\
    Pubmed45 &
    4.84 & 11.66 & 83.5 \\
    SPIDER &
    4.94 & 4.33 & 90.72 \\
    \bottomrule
    \end{tabular}
    \caption{Percentage of test samples that are helped (\% Helped), hurt (\% Hurt), or unchanged (\% Unchanged) when we change from \basemodel to \ourmodel using GPT-4.
    }
    \vspace{-1em}
    \label{tab:help_hurt_rate}
\end{table}

As shown in \cref{tab:help_hurt_rate}, AMR can change a significant proportion of answers, with 36.64\% changed on PAWS, and 37.62\% changed on WMT.
On its face, the lack of overall improvement from AMR supports the
current concern in the NLP community that traditional linguistics might have little role to play in improving the performance of NLP systems in the era of LLMs \cite{ignat-etal-2024-has}.
However, as there is a substantial subset of the data where AMR helps, if these improvements come from certain systematically identifiable subsets of the data, then this could provide clues for how structures such as AMR may potentially be leveraged to improve overall performance.
We investigate this question further 
in \cref{sec:q2,sec:q3}.

\begin{figure*}[!bhtp]

\begin{minipage}[t]{0.14\textwidth}
\begin{tikzpicture}
  \begin{axis}[
   width = 1.7\linewidth,
    title={PAWS},
     title style={font=\fontsize{8pt}{8pt}\selectfont},
    ylabel={F1},
    ylabel near ticks,  %
    ylabel shift=-3pt, %
    ybar,
    yticklabel style={inner sep=0.5pt},
    bar width=0.1cm,
    ymin=0,
    ymax=90,
    xtick=data,
    xticklabels={
      text-d.-001,
      text-d.-002,
      text-d.-003,
      gpt-3.5,  %
      gpt-4
    },
 xticklabel style={rotate=50, anchor=east, font=\fontsize{8pt}{8pt}\selectfont},
    yticklabel style={font=\fontsize{8pt}{8pt}\selectfont},
    ylabel style={font=\fontsize{8pt}{8pt}\selectfont},
    ymajorgrids=true,
    grid style=dashed,
    legend style={at={(0.02,0.98)}, anchor=north west, font=\footnotesize},
  ]
  \addplot+[ybar] coordinates {
    (1, 61.22)
    (2, 68.83)
    (3, 67.82)
    (4, 63.92)
    (5, 78.25)
  };

  \addplot+[ybar] coordinates {
    (1, 61.1)
    (2, 46.52)
    (3, 65.63)
    (4, 50.11)
    (5, 75.21)
  };
  \end{axis}
\end{tikzpicture}
\end{minipage}
\hspace{0.7cm} %
\begin{minipage}[t]{0.14\textwidth}
\begin{tikzpicture}
  \begin{axis}[
  width = 1.7\linewidth,
    title={WMT},
    title style={font=\fontsize{8pt}{8pt}\selectfont},
    ylabel={BLEU},
    ylabel near ticks,  %
    ylabel shift=-3pt, %
    ybar,
    yticklabel style={inner sep=0.5pt},
    bar width=0.1cm,
    ymin=0,
    ymax=40,
    xtick=data,
    xticklabels={
      text-d.-001,
      text-d.-002,
      text-d.-003,
      gpt-3.5,  %
      gpt-4,
    },
    xticklabel style={rotate=50, anchor=east, font=\fontsize{8pt}{8pt}\selectfont},
    yticklabel style={font=\fontsize{8pt}{8pt}\selectfont},
    ylabel style={font=\fontsize{8pt}{8pt}\selectfont},
    ymajorgrids=true,
    grid style=dashed,
    legend style={at={(0.02,0.98)}, anchor=north west, font=\footnotesize},
  ]
  \addplot+[ybar] coordinates {
    (1, 20.09)
    (2, 23.23)
    (3, 24.63)
    (4, 26.83)
    (5, 27.52)
  };
  \addplot+[ybar] coordinates {
    (1, 16.33)
    (2, 21.65)
    (3, 23.07)
    (4, 26.5)
    (5, 26.69)
  };
  \end{axis}
\end{tikzpicture}
\end{minipage}
\hspace{0.7cm} %
\begin{minipage}[t]{0.14\textwidth}
\begin{tikzpicture}
  \begin{axis}[
   width = 1.7\linewidth,
    title={Logic},
     title style={font=\fontsize{8pt}{8pt}\selectfont},
    ylabel={F1},
        ylabel near ticks,  %
    ylabel shift=-3pt, %
    ybar,
    yticklabel style={inner sep=0.5pt},
    bar width=0.1cm,
    ymin=0,
    ymax=60,
    xtick=data,
    xticklabels={
      text-d.-001,
      text-d.-002,
      text-d.-003,
      gpt-3.5,  %
      gpt-4
    },
    xticklabel style={rotate=50, anchor=east, font=\fontsize{8pt}{8pt}\selectfont},
    yticklabel style={font=\fontsize{8pt}{8pt}\selectfont},
    ylabel style={font=\fontsize{8pt}{8pt}\selectfont},
    ymajorgrids=true,
    grid style=dashed,
    legend style={at={(0.02,0.98)}, anchor=north west, font=\footnotesize},
  ]
  \addplot+[ybar] coordinates {
    (1, 23.52)
    (2, 44.83)
    (3, 41.98)
    (4, 43.77)
    (5, 55.61)
  };
  \addplot+[ybar] coordinates {
    (1, 18.05)
    (2, 38.14)
    (3, 37.57)
    (4, 43.77)
    (5, 55.12)
  };
  \end{axis}
\end{tikzpicture}
\end{minipage}
\hspace{0.7cm} %
\begin{minipage}[t]{0.14\textwidth}
\begin{tikzpicture}
  \begin{axis}[
   width = 1.7\linewidth,
    title={Pubmed45},
     title style={font=\fontsize{8pt}{8pt}\selectfont},
    ylabel={F1},
    ylabel near ticks,  %
    ylabel shift=-3pt, %
    ybar,
    yticklabel style={inner sep=0.5pt},
    bar width=0.1cm,
    ymin=0,
    ymax=60,
    xtick=data,
    xticklabels={
      text-d.-001,
      text-d.-002,
      text-d.-003,
      gpt-3.5,  %
      gpt-4
    },
 xticklabel style={rotate=50, anchor=east, font=\fontsize{8pt}{8pt}\selectfont},
    yticklabel style={font=\fontsize{8pt}{8pt}\selectfont},
    ylabel style={font=\fontsize{8pt}{8pt}\selectfont},
    ymajorgrids=true,
    grid style=dashed,
    legend style={at={(0.02,0.98)}, anchor=north west, font=\footnotesize},
  ]
  \addplot+[ybar] coordinates {
    (1, 24.1)
    (2, 26.93)
    (3, 27.96)
    (4, 30.53)
    (5, 39.65)

  };

  \addplot+[ybar] coordinates {
    (1, 24.66)
    (2, 25.84)
    (3, 26.2)
    (4, 28.87)
    (5, 35.78)
  };
  \end{axis}
\end{tikzpicture}
\end{minipage}
\hspace{0.7cm} %
\begin{minipage}[t]{0.14\textwidth}
\begin{tikzpicture}
  \begin{axis}[
   width = 1.7\linewidth,
    title={SPIDER},
     title style={font=\fontsize{8pt}{8pt}\selectfont},
    ylabel={Exact Match},
    ylabel near ticks,  %
    ylabel shift=-3pt, %
    ybar,
    yticklabel style={inner sep=0.5pt},
    bar width=0.1cm,
    ymin=0,
    ymax=60,
    xtick=data,
    xticklabels={
      text-d.-001,
      text-d.-002,
      text-d.-003,
      gpt-3.5,  %
      gpt-4
    },
 xticklabel style={rotate=50, anchor=east, font=\fontsize{8pt}{8pt}\selectfont},
    yticklabel style={font=\fontsize{8pt}{8pt}\selectfont},
    ylabel style={font=\footnotesize},
    ymajorgrids=true,
    grid style=dashed,
    legend style={at={(0.02,0.98)}, anchor=north west, font=\footnotesize},
  ]
  \addplot+[ybar] coordinates {
    (1, 9.27)
    (2, 35.86)
    (3, 33.85)
    (4, 38.25)
    (5, 43.78)
  };

  \addplot+[ybar] coordinates {
    (1, 1.98)
    (2, 28.31)
    (3, 34.97)
    (4, 35.86)
    (5, 44.39)
  };
  \end{axis}
\end{tikzpicture}
\end{minipage}
\caption{Performance of \basemodel (in purple) and \ourmodel (in red)  on 5 datasets across  5 model versions: text-davinci-001|-002|-003,  GPT-3.5 and GPT-4.} \label{fig:model_version}
\end{figure*}
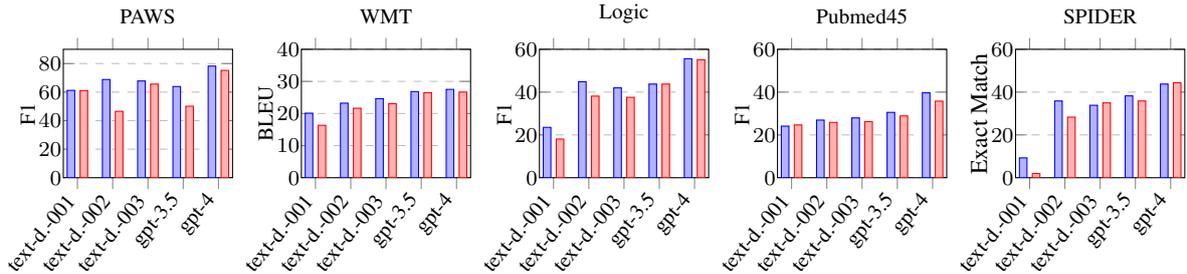
\subsection{AMR's Effect on Models with Different Capabilities}
\label{sec:model_version}

\cref{fig:model_version} shows the results of our experiments on models of varying capability,
from {text-davinci-001}, {-002}, {-003}, to GPT-3.5 and GPT-4.
Overall, \ourmodel hurts performance for most tasks and models, again with Text-to-SQL being the exception, at least for {text-davinci-003} and GPT-4. In some cases, less capable models degrade more when using AMR, which might be due to their limited ability to comprehend AMR and reason over its special symbols. This is consistent with our preliminary observations that none of the non--instruction-tuned earlier GPT models, or the less capable models such as LLaMa and Alpaca, comprehend AMR or reason over them well. 

\section{Q2: When Does AMR Help/Hurt?}\label{sec:q2}

The previous section shows that AMR is helpful or harmful for different samples.
Now we investigate the conditions under which it helps or harms performance, in particular whether this can be predicted from features of the input text.
We first illustrate a case study in \cref{sec:mwe}, where AMR's lack of ability to capture the semantic equivalence of multi-word expressions (MWEs) hinders paraphrase detection. Then, we perform two systematic interpretability studies: First, we treat linguistic features as our hypotheses, and extract features with high correlation with AMR helpfulness~(\cref{sec:linguistic}); second, we directly train classifiers to learn AMR helpfulness~(\cref{sec:classify}).

\subsection{Case Study: AMR's Shortcomings on MWEs}\label{sec:mwe}
AMR has its unique advantages and limitations, from which we can interpret what cases it can help, and what cases not. One such limitation of AMR is its lack of ability to capture MWEs such as idiomatic expressions, which makes it overlook certain semantic equivalences for paraphrase detection.
Consider the example in \cref{tab:slangeg}. Here, the proper paraphrase for the MWE \textit{swan song} is not ``bird song,'' but ``final performance.''
However, the AMRs for the three sentences do not reflect this; the AMR for the ``swan song'' sentence is structurally and lexically more similar to the ``bird song'' AMR than the one for the ``final performance'' variant.
\begin{figure}[ht]
\hspace*{0.4cm}
    \centering \small
    \setlength\tabcolsep{-15pt}
    \begin{tabular}{p{5cm}p{5cm}}
\textbf{Original Sentence with MWE} \\
\textit{Her \textbf{swan song} disappointed her fans.} \\
(z0 / disappoint-01 \newline
\hspace*{0.3em} :ARG0 (z1 / \textbf{song} \newline
\hspace*{0.3em} \hspace*{0.3em} :mod (z2 / \textbf{swan}) \newline
\hspace*{0.3em} \hspace*{0.3em} :poss (z3 / she)) \newline
\hspace*{0.3em} :ARG1 (z4 / fan \newline
\hspace*{0.3em} \hspace*{0.3em} :poss z3))
\\
\\
\textbf{Paraphrase Candidate 1}  & \textbf{Paraphrase Candidate 2} \\
    (\xmark Not a paraphrase.) & (\cmark A paraphrase.)
    \\
    \\
\textit{Her \textbf{bird song} disappointed \newline her fans.}
&\textit{Her \textbf{final performance}  \newline disappointed her fans.}
\\
(z0 / disappoint-01 \newline
\hspace*{0.3em} :ARG0 (z1 / \textbf{song} \newline
\hspace*{0.3em} \hspace*{0.3em} :mod (z2 / \textbf{bird}) \newline
\hspace*{0.3em} \hspace*{0.3em} :poss (z3 / she)) \newline
\hspace*{0.3em} :ARG1 (z4 / fan \newline
\hspace*{0.3em} \hspace*{0.3em} :poss z3))
    &
(z0 / disappoint-01\newline
\hspace*{0.3em} :ARG0 (z1 / \textbf{perform}-02\newline
\hspace*{0.3em} \hspace*{0.3em} :ARG0 (z2 / she)\newline
\hspace*{0.3em} \hspace*{0.3em} :mod (z3 / \textbf{final}))\newline
\hspace*{0.3em} :ARG1 (z4 / fan\newline
\hspace*{0.3em} \hspace*{0.3em} :poss z2))
    \end{tabular}
    \caption{
    An example showing the failure of AMR for paraphrase detection when the original sentence involves a MWE. This example is from our GoldSlang-ComposedAMR dataset.
    }
    \vspace{-1em}
    \label{tab:slangeg}
\end{figure}

Given this intuition, we qualitatively study whether AMR systematically fails on texts that contain more MWEs. %
We run \ourmodel on a self-composed dataset of paraphrase detection involving slang, assuming slang has more MWEs.
Since our experiments need annotations for both slang paraphrase pairs and AMRs, we compose two datasets, GoldSlang-ComposedAMR, and GoldAMR-ComposedSlang.
For GoldSlang-ComposedAMR, we use the curated slang paraphrase pairs by \citet{tayyar2021astitchinlanguagemodels} and generate their AMRs with an off-the-shelf parser~\citep{drozdov2022inducing}.
For GoldAMR-ComposedSlang, we use gold AMRs from the LDC AMR 3.0 corpus \cite{banarescu2013abstract}, and compose slang paraphrases using a combination of manual annotation and assistance from GPT-4. The data curation steps and data statistics are in \cref{appd:slang_annot}.

\begin{table}[b]
    \centering
    \small
    \setlength\tabcolsep{6pt}
    \begin{tabular}{llccccccc}
    \toprule
    Dataset
    & \basemodel & $\Delta$\ourmodel
    \\ \midrule

GoldSlang-ComposedAMR
 &
 86.83 & -6.63\\

GoldAMR-ComposedSlang & 77.69 & -8.78\\

    \bottomrule
    \end{tabular}
    \caption{\ourmodel results in a large drop in performance on slang-comprising paraphrase detection data. 
    }
    \label{tab:slang_res}
\end{table}
\cref{tab:slang_res} shows evaluation results, where \ourmodel causes a large drop in performance compared to \basemodel, more substantial than the slight fluctuation of -3 to +1 percentage points shown previously in \cref{tab:main_res}. It is very likely that, due to the shortcomings of AMR on MWEs, \ourmodel mostly distracts the model, yielding worse performance.

\subsection{Large-Scale Text Feature Analysis}\label{sec:linguistic}

The case study above provides a precise insight into a special case when AMR does not work.
To systematically explore a larger set of hypotheses, we perform a feature analysis over the input texts.
We formulate the contribution of AMR as the \textit{AMR helpfulness} score, which is the per-example performance difference between \ourmodel and \basemodel, ranging between -100\% and 100\%, where a negative value means that AMR hurts performance on the example, and a positive value means that AMR improves performance.

For each input,
we compute a comprehensive set of linguistic features, including 139
features on the text representation, and 4 features derived from the AMR.
Specifically, we obtain 55 features using the Text Characterization Toolkit (TCT)
\citep{simig2022text}, which is specifically designed to facilitate the analysis of text dataset properties, 17 different part-of-speech (POS) tags,
44 dependency tags,
and 61 other hand-crafted features, which characterize the semantic and syntactic complexity of the input text, such as the {number of arguments vs.~adjuncts}~\citep{haspelmath2014Arguments}.

\begin{table}[t]
    \centering \small
    \begin{tabular}{lc}
    \toprule\
    Top 5 Positive Features & Pearson Correlation \\ 
    \midrule
    \textit{Adj} POS Tag Frequency & 0.0393\\
    Avg. Word Complexity & 0.0343 \\
    \# Adjuncts & 0.0337 \\
    Max Word Complexity & 0.0316 \\
    Avg. Word Frequency & 0.0271\\
    \bottomrule
    \end{tabular}
    \caption{
    The top five features with the highest positive correlation coefficients to AMR helpfulness: the frequency of adjectives among all the words (\textit{Adj} POS Tag Frequency), average word complexity level by the age of acquisition \citep{Kuperman2012Ageofacquisition}, number of adjuncts, maximum word complexity level by the age of acquisition, and average word frequency.
}
    \label{tab:positive_corr_feature}
    \vspace{1em}
\end{table}

\cref{tab:positive_corr_feature,tab:negative_corr_feature} show the Pearson correlation between each linguistic feature and the AMR helpfulness score.
Overall, the correlation of each individual feature to the AMR helpfulness score is not strong, either because these features do not explain much about AMR helpfulness, or because it requires a combination of multiple features. %
Though the correlations are weak, the top correlated features in \cref{tab:positive_corr_feature} align with our intuition that AMR should be helpful for semantically complex sentences:
AMR is most helpful for samples with more adjectives, complex words, and adjuncts. 
In
\cref{tab:negative_corr_feature}, the top negative feature, the number of named entities, echoes the finding in our previous MWE case study in \cref{sec:mwe}, and we systematically show that AMR
 is most harmful on samples with many named entities, tokens containing digits, and proper nouns.

\begin{table}[t]
    \centering \small
    \setlength\tabcolsep{3pt}
    \begin{tabular}{lc}
    \toprule
    Top 5 Negative Features & Pearson Correlation \\ 
    \midrule
    \# Named Entities & -0.0630 \\
    \% of Tokens Containing Digits & -0.0281 \\
    \# Proper Nouns & -0.0258 \\
    \# Third Person Singular Pronouns & -0.0236 \\
    \# Quantifier Phrase Modifiers & -0.0222 \\
    \bottomrule
    \end{tabular}
    \caption{The top five features with the highest negative correlation coefficients to AMR helpfulness: the number of named entities, percentage of tokens containing digits, number of proper nouns (e.g., London), number of third person singular pronouns (e.g., he), and number of quantifier phrase modifiers. See detailed explanations of features in \cref{sec:feature_explanation}.}
    \label{tab:negative_corr_feature}
\end{table}

\subsection{AMR Helpfulness Prediction as a Learning Task
}\label{sec:classify}
Now we analyze the upper-bound predictability of AMR helpfulness from the input, both on the basis of our linguistic features and text input itself.
Specifically, we train models to predict AMR helpfulness as a binary classification task where the positive class is the case where AMR helps, and the negative class is the rest.
Merging all five datasets together, we have a binary classification dataset of 19,405 training samples, 4,267 development samples, and 5,766 test samples, with positive labels composing 10.38\% of the dataset.

\begin{table}[b]
\centering \small
\begin{tabular}{lccccc}
\toprule
Model & F1 & Acc & P & R\\\midrule
Random Baseline & 16.14 & 49.95 & 9.65 & 49.16 \\ \hline %
\multicolumn{5}{l}{\textit{Using Linguistic Features}}\\
Random Forest& \textbf{32.67} & 81.93 & 25.72 & 44.75\\
XGBoost& 30.08 & 78.47 & 22.06 & 47.27\\ 
Ensemble& 30.42 & 77.59 & 21.85 & 50.00\\ \hline
\multicolumn{5}{l}{\textit{Using the Free-Form Text Input}}\\
{BERT} & 
\textbf{33.83}
& 79.70
& 25.00
& 52.28

\\
{RoBERTa} & 33.29
& 80.36
& 25.11
& 49.38 \\
\bottomrule
\end{tabular}
\caption{Classification performance of various models on AMR helpfulness. We report the F1, precision (P), and recall (R) of the positive class, as well as the accuracy (Acc). See the implementation details of the models in \cref{app:impl}.
}
\label{tab:amr_helpfulness_prediction}
\end{table}

As shown in \cref{tab:amr_helpfulness_prediction}, classifiers based on linguistic features achieve an F1 score of up to {32.67}\%. BERT-based deep learning models improve by up to 1.16 F1 scores, with substantial increases in recall.
For interpretability, we run Shapley feature attribution method~\citep{Fryer2021Shapley} and find that words that signal the existence of clauses tend to have high importance for the classifier, such as ``what,'' ``how,'' ``said,'' and ``says.'' These results do not provide a clear explanation of when AMR can help, but give a starting point, and we welcome future research to continue exploring the potential benefits of AMR. The fact that AMR helpfulness is challenging to predict even for BERT models may indicate either that we need more data to learn the features that predict this, or that a substantial portion of the changes that AMR makes to model predictions correspond to noise (i.e., help or hurt in unpredictable ways).

\section{Q3: Why Does AMR Help/Hurt?}\label{sec:q3}
To understand why AMR helps or hurts when it does,
we look into the following subquestions: (1) how does parser-generated AMR work compared with gold AMR (\cref{sec:gold})? (2) what is the representation power of AMR versus text when the other is ablated (\cref{sec:ablation})? And (3) how does AMR help in each step of the reasoning process (\cref{sec:step})?

\subsection{Gold vs. Parser-Generated AMR}\label{sec:gold}

First, we investigate whether there are cascading errors before the CoT process, due to mistakes in the parser-generated AMR. For example, the reported performance of \citet{drozdov2022inducing} is 83\% on AMR 3.0 \cite{banarescu2013abstract}. To assess this, we compare \ourmodel performance when using predicted versus gold AMRs.
Testing this requires data with gold AMR annotations as well as gold labels for some downstream NLP task we can evaluate the models on.
To this end, we take the intersection of the AMR 3.0 dataset~\citep{banarescu2013abstract} with Ontonotes 5.0~\citep{pradhan2011conll},
which contains 131 sentences that have both gold AMR and named entity recognition (NER) annotations. We list the intuition of why AMR can be helpful for NER in \cref{appd:intuition}.

\begin{table}[b]
    \centering
    \small
    \begin{tabular}{cccccccccccc}
    \toprule
    Dataset
    & \basemodel & AMR  & $\Delta$\ourmodel
    
    \\
\midrule
\multirow{2}{*}{AMR-NER} & \multirow{2}{*}{60.51} & Gold & +0.03
\\
& & Parser & +1.91
\\

    \bottomrule
    \end{tabular}
    \caption{Model performance on the AMR-NER data using the gold AMR (Gold) and parser-generated AMR (Parser).
    We report the \basemodel performance, and the change of performance by \ourmodel ($\Delta$\ourmodel) in terms of F1 scores. 
    }
    \label{tab:amr_ner}
\end{table}

Using this AMR-NER dataset, we compare the performance of \ourmodel with gold AMR versus parser-generated AMR on NER,
shown in \cref{tab:amr_ner}. Both lead to similar results, with a difference of less than two percentage points (which is not statistically significant, with $p$ = 0.627 by t-test). The test set is unfortunately too small to reliably detect an effect of reasonable size, due to the lack of available data with both gold AMR and NLP task annotations; this result is also specific to NER, which may not have all of the relevant features for understanding the effect of gold versus automatically produced AMRs. However, the fact that the observed effect size is very small constitutes some evidence that improving the predicted AMRs would likely not play a huge role in increasing downstream performance with current models.

\subsection{Ablating the AMR/Text Representation
}\label{sec:ablation}
As discussed in \cref{sec:repr_power}, AMR and text representations are two different surface forms for expressing sentence semantics, but one representation may be more useful to the LLM than the other.
To test this, we conduct an ablation study removing either the original text or the AMR and measuring performance (see \cref{app:ablation}).
To avoid the potential for cascading errors from the parsing process, we use the AMR-NER dataset with the gold AMR.
\begin{figure}[ht]
\centering
\begin{tikzpicture}
\begin{axis}[
    xlabel={
    AMR/Text Tokens (\%) Ablated in the Prompt},
    ylabel={Task Performance (\%)},
    xmin=0, xmax=100,
    ymin=30, ymax=70,
    ymajorgrids=true,
    grid style=dashed,
    width=0.6\columnwidth,
    tick label style={font=\footnotesize},
    legend style={at={(1.04,0.2)}, anchor=west, font=\footnotesize},
    legend style={font=\footnotesize, cells={align=left}},
    label style={font=\footnotesize}
]

\addplot[
    color=blue,
    mark=square,
]
coordinates {

    (100,53.68)
    (80, 58.88)
    (60,61.24)
    (40,56.18)
    (20,58.52)
    (0,60.54)

}; %
\addlegendentry{\% AMR Ablated}

\addplot[
    color=orange,
    mark=*,
]
coordinates {

    (100,34.63)
    (80,39.43)
    (60,39.71)
    (40,42.17)
    (20,47.95)
    (0,60.54)

};
\addlegendentry{\% Text Ablated}

\end{axis}
\end{tikzpicture}
    \caption{Ablation studies of AMR and text representations in the prompt on the AMR-NER dataset using {GPT-4}.
    Starting from the \ourmodel prompt with the complete text and AMR, we randomly drop out a certain portion of tokens in the text/AMR, and see the effect on the task performance. 
    }
    \vspace{-1em}
    \label{fig:amr_text_ablation}
\end{figure}

\paragraph{Results}
In addition to previous results contrasting \ourmodel, which provides both the text and AMR in the prompt, and \basemodel with the text-only input, we
show the results of a more granular analysis in \cref{fig:amr_text_ablation}, where we randomly drop out text and AMR tokens and measure the effect on task performance. Similar to the above, we find that dropping AMR marginally decreases performance, and dropping text much more drastically degrades LLM performance, showing the greater utility of text as a representation for LLMs. We also conduct the same ablation study on 1,000 random samples from the WMT dataset using predicted AMRs in \cref{appd:ablation}, where the observations are similar.

\subsection{Checking the Step-By-Step Reasoning}
\label{sec:step}
To better understand how LLMs use AMR, we directly examine the step-by-step reasoning process produced by \ourmodel with GPT-4.
We randomly select 50 samples from the PAWS dataset and manually annotate the correctness of each step in the reasoning process.
For paraphrasing on PAWS, the steps (and our evaluations) are as follows:
\begin{enumerate}
\item Produce the AMR for the input sentences using \citet{drozdov2022inducing}'s structured BART model. Instead of manually annotating correctness of these AMRs, we defer to their reported performance of 82.6 SMATCH scores on the AMR 3.0 dataset.
\item Provide the AMRs to GPT-4 in the paraphrasing task prompt using \ourmodel, and then instruct it to list all the commonalities and differences of the AMRs. Our manual check finds that GPT-4 achieves a 97\% F1 score (with 95\% precision, 98\% recall) at listing these.
\item GPT-4 then outputs a final decision on whether the sentences are paraphrases. We evaluate that its judgment in this step is consistent with the reasoning in the prior step 80\% of the time.
\end{enumerate}

Even though GPT-4 was able to correctly enumerate the relevant features of the AMRs, it still had trouble synthesizing this information into a correct paraphrasing judgment.
These mistakes as well as the potential for cascading errors may explain why \ourmodel achieves a performance of 75.21\% on PAWS, which is a slight drop from the \basemodel performance of 78.25\%.
Overall, this provides further evidence of the advantages that free-form text itself has as a representation for LLMs to operate on.

\section{Related  Work}
\paragraph{Semantic Representations}
Traditionally, NLP models often represent text by features developed on the basis of linguistic expertise, among which semantic representations such as AMR \cite{banarescu2013abstract} are used to abstract away the surface form of the text and distill the most important elements of its meaning. In the past, such representations have helped with a variety of NLP tasks, such as
semantic parsing \citep{kuhn1995the}, machine translation \citep{wu2009semantic,wong2006learning}, and text summarization \citep{liu2018toward}. 
Recent research also looks into whether LLMs already incorporate a good understanding of semantic representations
\cite[e.g.,][]{staliunaite-iacobacci-2020-compositional,blevins-etal-2023-prompting}.

\myparagraph{Chain-of-Thought Prompting}
Rapid advancement in LLMs has led to a new paradigm of
performing NLP tasks by eliciting model behavior via instructions and examples using prompting \cite{brown2020gpt3,raffel2020exploring}.
Chain-of-thought (CoT) prompting ~\citep{wei2022chain,nye2021show}, which pairs input examples with step-by-step explanations of how to produce their respective outputs, has been shown to improve LLMs' performance at 
various reasoning tasks~\citep{yu-etal-2023-alert}, and its variants have also shown success in various scenarios, such as \textsc{CausalCoT} for causal reasoning \cite{jin2023cladder}, and \textsc{MoralCoT} for moral reasoning \cite{jin2022make}. 

Our work proposes a way to bridge linguistic representations with text by \ourmodel, 
providing an AMR as an intermediate representation for the LLM to reason over.
Our results are mixed, demonstrating the relative advantage that unstructured, free-text representations have for language models pretrained on large amounts of natural language data.

\section{Conclusion}
In this work, we analyze the role of semantic representations in the era of LLMs. In response to the ongoing paradigm shift in the NLP community, we show that AMR in general is not yet a representation immediately fit for pre-trained LLMs. However, our study shows that AMR still helps on some samples.
We also suggest that a potential direction to enhance AMR's contribution to LLMs is to improve the understanding of LLMs over the schemes and symbols of AMR, and map it to the reasoning of the respective NLP task. This work presents an effort to bridge the traditionally rich linguistic structures with the strength of LLMs.
\section*{Limitations and Future Work}
This work explores one form of linguistic representation of text. In the future, we welcome more exploration on various other linguistic representations using the methodology presented in this work. Moreover, we explore one intuitive way of prompting the model.
Future work is welcome to explore different ways of prompting to make the AMR information more accessible and useful to the model.

In addition, some of our analyses are limited by a lack of annotated resources, so we were only able to show experimental results on hundreds of examples in some cases where gold AMR annotation is needed. This is a commonly known issue for AMR, which is expensive and requires a high level of linguistic expertise to annotate. This limitation makes the results less statistically significant than what we could have if there are more annotated AMRs available.
In this work, we hope to strike a balance to still show some meaningful trends while trying to get the largest size of annotated data we can.

Moreover, if any future work has the resources to train an LLM specifically optimized for AMR as a representation, this would be the ideal setting to check out the upper bound of the power of AMR in the era of LLMs.%

As for the limitations for specific parts of the paper, for example for the notion of gold AMRs in \cref{sec:q3}, although we use the AMR annotated by humans in the official \citet{banarescu2013abstract} dataset, it should be noted that such AMRs are not necessarily ``perfect'', as humans might also have a non-perfect inter-annotator agreement over some AMRs. And while SMATCH scores can be predictive, they may not perfectly reflect the quality of parser-generated AMRs \cite{opitz-frank-2022-better}. These are both open research questions, and we use the AMRs released by the official source \cite{banarescu2013abstract} as a proxy for ground-truth AMR.

\section*{Ethical Considerations}
The datasets used in this paper are existing public datasets on general NLP tasks without any user-sensitive information. We are not aware of specific ethical concerns with the analysis in this study, which is a neutral investigation to understand the role of traditional linguistic structures such as semantic representations in the era of LLMs.

\ifarxiv

\section*{Acknowledgments}

We thank Juri Opitz for his insightful suggestions on our AMR experiments based on profound domain expertise.
We also appreciate Wendong Liang for insightful discussions on Komolgorov complexity, which is a foundation of the theoretical framework in this work.
We thank Nils Heil for extracting the SQL schemes of the SPIDER dataset so that we can incorporate them in the prompt to improve our performance. 
This material is based in part upon works supported by the German Federal Ministry of Education and Research (BMBF): Tübingen AI Center, FKZ: 01IS18039B; and by the Machine Learning Cluster of Excellence, EXC number 2064/1 – Project number 390727645;
Zhijing Jin is supported by PhD fellowships from the Future of Life Institute and Open Philanthropy.
\section*{Author Contributions}\label{sec:contributions}
Mona Diab initiated the project idea based on her strong expertise in traditional linguistics, and an intuition that the semantic representations should help model efficiency, robustness, and interpretability. During the course of exploration by Zhijing Jin and Mona Diab together for over a year, they find that the AMR representations does not always help LLMs over multiple experimental setups and model implementations. 

Zhijing further explores the theoretical formulation of representation power to provide the explanations behind the observed performance, together with the expertise of Bernhard Schoelkopf in causal representation learning.
Julian Michael provided valuable insights and overview of the field of semantic representations, which brings the depth of the project to another level. Julian also
provided constructive suggestions for improving the experiments and structuring the paper, and substantially improved the writing. 

Yuen Chen and Fernando Gonzalez contributed substantially to scaling up all the experiments across multiple datasets and multiple model versions, and analyzing the results. Jiarui Liu and Jiayi Zhang helped with the training the BERT-based classifiers, and analyzing the Shapley values. Jiarui Liu conducted several important experiments for the camera-ready version of the paper, especially on checking the ceiling performance of \ourmodel with various prompt improvements and data setups.

\fi
\bibliography{sec/refs_zhijing,sec/refs_causality,ref}
\bibliographystyle{acl_natbib}

\cleardoublepage

\appendix
\section{Experimental Details}

\subsection{Data Split Details}
\label{appd:split}
For the five datasets that we use in \cref{sec:q2}, we compose the test sets in the following way. To make the experimental results in our paper representative, we aim at composing a large test set for each task, ideally more than a size of 5,000 test samples. We sequentially check whether the test set is large enough, and if not, then we include the development set and the training set sequentially. Note that since our experiments are zero-shot, i.e., we do not train our models at all, any of the original test, development, or training sets can be used to report the performance on.

As a result, for the PAWS dataset, we use its entire test set, which is large enough with 8,000 samples.
For WMT16, since its test set has only 2,999 samples, we also include its development set with 3,000 samples, totaling 5,999 samples for our test.
For the LOGIC dataset, as it is a relatively small dataset, we add up all its original test, development, or training sets to obtain 2,449 samples.
For Pubmed45, which contains 25,360 unsplit samples, we randomly select 5,000 data points for our analysis. 
For SPIDER, as its test set is not released, and development set has only 1,034 samples, we also include its training set of 7,000 samples, totaling 8,034 samples for our test.

\subsection{Evaluation Metrics}\label{appd:eval}
For evaluation, we report the performance of PAWS, Logic, and Pubmed45 by F1 scores, the performance of machine translation on the WMT16 dataset
by BLEU scores \citep{papineni2002bleu}, and the performance of text-to-SQL generation using the official evaluation setup at \url{https://github.com/taoyds/test-suite-sql-eval}.
To evaluate the generation quality of parser-produced AMRs, we report the SMATCH scores using the evaluation codes at \url{https://github.com/snowblink14/smatch}.

\subsection{Prompts}\label{appd:prompts}
We list the prompts for \basemodel and \ourmodel of all datasets in \cref{tab:prompts,tab:prompts2}, as well as the system prompts in \cref{tab:system_prompts}.
\begin{table}[ht]
    \centering \tiny
    \begin{tabular}{lp{6cm}}
\toprule
\multicolumn{2}{l}{\textbf{\textit{Paraphrase Detection (PAWS)}}} \\
\basemodel & Paraphrase Detection: Determine if the following two sentences are exact paraphrases (rewritten versions with the same meaning) of each other.\newline Sentence 1: \texttt{\{sentence1\}}\newline Sentence 2: \texttt{\{sentence2\}}\newline Answer [Yes/No] and then provide a brief explanation of why you think the sentences are paraphrases or not.\newline Paraphrase:
\\
\ourmodel & Paraphrase Detection: You are given two sentences and the abstract meaning representation (AMR) of each.\newline Sentence 1: \texttt{\{sentence1\}}\newline AMR 1:\newline \texttt{\{amr1\}}\newline Sentence 2: \texttt{\{sentence2\}}\newline AMR 2:\newline \texttt{\{amr2\}}\newline Explain what are the commonalities and differences between the two AMRs. Then determine if the two sentences are exact paraphrases (rewritten versions with the same meaning) of each other and provide a brief explanation of why you think the sentences are paraphrases or not. Use the following format: Answer: [Yes/No]
\\
\midrule
\multicolumn{2}{l}{\textbf{\textit{Translation (WMT16)}}}
\\ \basemodel & Please translate the following text from English to German.\newline Text:  \texttt{\{sentence1\}}\newline Translation:
\\
\ourmodel & You are given a text and its abstract meaning representation (AMR).\newline Text: \texttt{\{sentence1\}}\newline AMR:\newline \texttt{\{amr1\}}\newline Please translate the text from English to German. You can refer to the provided AMR if it helps you in creating the translation.\newline Translation:
\\
\midrule
\multicolumn{2}{l}{\textbf{\textit{Logical Fallacy Detection (Logic)}}}
\\ \basemodel & Text: \texttt{\{sentence1\}}.\newline You must answer one option from the listed categories without additional text.
\\
\ourmodel & Text: \texttt{\{sentence1\}}\newline AMR:\newline \texttt{\{amr1\}}\newline You must answer one option from the listed categories without additional text.
\\
\midrule
\multicolumn{2}{l}{\textbf{\textit{Event Extraction (Pubmed45)}}}
\\ \basemodel & This question aims to assess your proficiency in validating relationships between different entities in biomedical text. You will be presented with a sentence from an article and asked to determine whether the interaction between the entities mentioned in the sentence is valid or not. You should respond with a single digit, either "0" if the interaction is invalid, "1" if it is valid, or "2" if swapping the positions of any two entities would make the interaction valid. Please note that you are required to provide only one of these three responses.\newline Text:  \texttt{\{sentence1\}}\newline Interaction: \texttt{\{interaction\}}
\\
\ourmodel & This question aims to assess your proficiency in validating relationships between different entities in biomedical text. You will be presented with a sentence from an article and its abstract meaning representation (AMR) and asked to determine whether the interaction between the entities mentioned in the sentence is valid or not. You should respond with a single digit, either "0" if the interaction is invalid, "1" if it is valid, or "2" if swapping the positions of any two entities would make the interaction valid. Please note that you are required to provide only one of these three responses.\newline Text: \texttt{\{sentence1\}}\newline AMR:\newline \texttt{\{amr1\}}\newline Interaction: \texttt{\{interaction\}}
\\
\bottomrule
    \end{tabular}
    \caption{Prompts for 
    PAWS, WMT16, Logic, and Pubmed45.}
    \label{tab:prompts}
\end{table}

\begin{table}[ht]
    \centering \tiny
    \begin{tabular}{lp{6cm}}
\toprule
\multicolumn{2}{l}{\textbf{\textit{Text2SQL (SPIDER)}}}
\\ \basemodel & Write an SQL query that retrieves the requested information based on the given natural language question. Remember to use proper SQL syntax and consider any necessary table joins or conditions.\newline Question: \texttt{\{sentence1\}}\newline Query:
\\
\ourmodel & Write an SQL query that retrieves the requested information based on the given natural language question and its abstract meaning representation (AMR). Remember to use proper SQL syntax and consider any necessary table joins or conditions.\newline Question: \texttt{\{sentence1\}}\newline AMR:\newline \texttt{\{amr1\}}\newline Query:
\\
\midrule
\multicolumn{2}{l}{\textbf{\textit{Named Entity Recognition (AMR-NER)}}}
\\ \basemodel & The following is a named entity recognition task. Please extract all the named entities of the following types from the given sentence.
TYPE="CARDINAL": Numerals that do not fall under another type, e.g., “one”, “ten”
TYPE="DATE": Absolute or relative dates or periods. E.g., “the summer of 2005”, “recent years”
TYPE="EVENT": Named hurricanes, battles, wars, sports events, etc. E.g., “Olympiad games”
TYPE="FAC": Buildings, airports, highways, bridges, etc. E.g., “Disney”, “the North Pole”
TYPE="GPE": Countries, cities, states. E.g., “Hong Kong”, “Putian”
TYPE="LAW": Named documents made into laws. E.g., “Chapter 11 of the federal Bankruptcy Code”
TYPE="LOC": Non-GPE locations, mountain ranges, bodies of water. E.g., “Mai Po Marshes”, “Asia”
TYPE="MONEY": Monetary values, including unit. E.g., “\$ 1.3 million”, “more than \$ 500 million”
TYPE="NORP": Nationalities or religious or political groups. E.g., “Chinese”, “Buddhism”
TYPE="ORDINAL": E.g., "first", "second", etc.
TYPE="ORG": Companies, agencies, institutions, etc. E.g., “Eighth Route Army”, “the Chinese Communist Party”
TYPE="PERCENT": Percentage, including "\%". E.g., “25 \%”
TYPE="PERSON": People, including fictional. E.g., “Zhu De”, “Saddam Hussein”
TYPE="PRODUCT":  Objects, vehicles, foods, etc. (Not services.) E.g., “iPhone”, “Coke Cola”
TYPE="QUANTITY": Measurements, as of weight or distance. E.g., “23 sq. km”
TYPE="TIME": Times smaller than a day. E.g., “homecoming night”
Sentence:  \texttt{\{sentence1\}}\newline Use json format for the response where each key is an entity type.
\\
\ourmodel & The following is a named entity recognition task. Please extract all the named entities of the following types from the given sentence and its abstract meaning representation (AMR).
TYPE="CARDINAL": Numerals that do not fall under another type, e.g., “one”, “ten”
TYPE="DATE": Absolute or relative dates or periods. E.g., “the summer of 2005”, “recent years”
TYPE="EVENT": Named hurricanes, battles, wars, sports events, etc. E.g., “Olympiad games”
TYPE="FAC": Buildings, airports, highways, bridges, etc. E.g., “Disney”, “the North Pole”
TYPE="GPE": Countries, cities, states. E.g., “Hong Kong”, “Putian”
TYPE="LAW": Named documents made into laws. E.g., “Chapter 11 of the federal Bankruptcy Code”
TYPE="LOC": Non-GPE locations, mountain ranges, bodies of water. E.g., “Mai Po Marshes”, “Asia”
TYPE="MONEY": Monetary values, including unit. E.g., “\$ 1.3 million”, “more than \$ 500 million”
TYPE="NORP": Nationalities or religious or political groups. E.g., “Chinese”, “Buddhism”
TYPE="ORDINAL": E.g., "first", "second", etc.
TYPE="ORG": Companies, agencies, institutions, etc. E.g., “Eighth Route Army”, “the Chinese Communist Party”
TYPE="PERCENT": Percentage, including "\%". E.g., “25 \%”
TYPE="PERSON": People, including fictional. E.g., “Zhu De”, “Saddam Hussein”
TYPE="PRODUCT":  Objects, vehicles, foods, etc. (Not services.) E.g., “iPhone”, “Coke Cola”
TYPE="QUANTITY": Measurements, as of weight or distance. E.g., “23 sq. km”
TYPE="TIME": Times smaller than a day. E.g., “homecoming night”
Sentence: \texttt{\{sentence1\}}\newline AMR:\newline \texttt{\{amr1\}}\newline Use json format for the response where each key is an entity type.
\\
\bottomrule
    \end{tabular}
    \caption{Prompts for 
SPIDER and AMR-NER.}
    \label{tab:prompts2}
\end{table}

\begin{table}[ht]
    \centering \small\setlength\tabcolsep{2pt}
    \begin{tabular}{lp{6.2cm}}
\toprule
PAWS & You are an NLP assistant whose purpose is to perform Paraphrase Identification. The goal of Paraphrase Identification is to determine whether a pair of sentences have the same meaning. \\ 
\hline
WMT16 & You are an NLP assistant expert in machine translation from English to German. \\
\hline
Logic & You are an expert in logic whose purpose is to determine the type of logical fallacy presented in a text or complete the text with one of the following logical fallacies.  1) Faulty Generalization\newline 2) False Causality\newline 3) Circular Claim\newline 4) Ad Populum\newline 5) Ad Hominem\newline 6) Deductive Fallacy\newline 7) Appeal to Emotion\newline 8) False Dilemma\newline 9) Equivocation\newline 10) Fallacy of Extension\newline 11) Fallacy of Relevance\newline 12) Fallacy of Credibility\newline 13) Intentional Fallacy. \\
\hline
Pubmed45 & You are a medical professional expert. \\
\hline
SPIDER & You are a language model designed to generate SQL queries based on natural language questions. Given a question, you need to generate the corresponding SQL query that retrieves the requested information from a database. \\
\hline
AMR-NER & You are an NLP assistant whose purpose is to perform named entity recognition (NER). \\
\bottomrule
    \end{tabular}
    \caption{System prompts for all datasets.}
    \label{tab:system_prompts}
\end{table}

\subsection{Example Data Samples}
To get a better sense of how the data samples look, we provide some example (text, AMR) pairs in \cref{tab:eg_text_amr}. 

\subsection{Implementation Details}\label{app:impl}

As for the experimental details, for the BERT and RoBERTa models, we use the weighted cross entropy loss, with a batch size of 16, learning rate of 1e-5, and dropout of 0.1, and train for five epochs until convergence.
For the XGBoost classifier \cite{chen2016xgboost}, we use the default hyperparameters, and set the random seed to 0, and the class weight proportional to the class ratio, namely setting the positive weight to be the inverse of the number of samples in the positive class divided by that of the negative class.

\subsection{Details of Ablation Study}\label{app:ablation}
For text/AMR ablation experiments, we use \ourmodel prompt with portions of text/AMR string ablated. An example of ablating 100\% of the AMR is as follows: 

\textit{``You are given a text and its abstract meaning representation (AMR).\\
Text: The relationship between Obama and Netanyahu is not exactly friendly.\\
AMR:\\
Please translate the text from English to German. You can refer to the provided AMR if it helps you in creating the translation.
Translation:''}

We also provide an example of ablating 100\% of the text:

\textit{``You are given a text and its abstract meaning representation (AMR).\\
Text: \\
AMR:\\
(f / friendly-01~9\\
    :ARG1 (r / relation-03~1\\
        :ARG0 (p / person~3\\
            :name (n / name~3\\
                :op1 "Obama"~3))\\
        :ARG2 (p2 / person~5\\
            :name (n2 / name~5\\
                :op1 "Netanyahu"~5)))\\
    :mod (e / exact~8)\\
    :polarity -~7)\\
Please translate the text from English to German. You can refer to the provided AMR if it helps you in creating the translation.
Translation:''}
\section{Data Collection}
\subsection{Composing the Slang-Involved Paraphrase Detection Dataset}\label{appd:slang_annot}

Since our experiments need annotations for both slang paraphrase pairs and AMRs, we compose two datasets, GoldSlang-ComposedAMR, and GoldAMR-ComposedSlang.
For GoldSlang-ComposedAMR, we use the curated slang paraphrase pairs by \citet{tayyar2021astitchinlanguagemodels}, and generate their AMRs with an off-the-shelf parser \cite{drozdov2022inducing}.
For the other dataset, GoldAMR-ComposedSlang, we use gold AMRs from the LDC AMR 3.0 corpus \cite{banarescu2013abstract}, and compose slang paraphrases using a combination of human efforts and assistance from GPT-4.

\paragraph{Composing the GoldSlang-ComposedAMR Dataset}

We adapt a subset of the ASILM \cite{tayyar2021astitchinlanguagemodels}, an idiomatic MWE dataset, into a paraphrase detection task. Each sentence in the subset containing idiomatic expressions is paired with a paraphrase (where the idiom is replaced with its literal semantic equivalent) and a non-paraphrase (where the idiom is replaced with a phrase of similar superficial meaning but differing semantic meaning). This results in a balanced paraphrase detection dataset with respect to ground truth labels.

\paragraph{Composing the GoldAMR-ComposedSlang Dataset}

A possible error in \ourmodel lies in the imperfection of parser-generated AMRs.
To disentangle the harm caused by (1) incorrect AMRs produced by the parsers and (2) poor representation of slang expressions by AMRs, we handcrafted the GoldAMR-Slang-Para dataset. We first extract a subset from LDC-AMR3.0 \citep{banarescu2013abstract} that involve slang expressions. Then, for each sentence, we replace the slang expression with an alternative expression of the same meaning, and a semantically different expression which seems literally similar, thus creating a paraphrase and non-paraphrase sentence, respectively. The corresponding AMRs can be derived from the original LDC-AMR3.0 AMRs with minimal modifications.

Specifically, we operationalize the process as follows.
We first use gpt-3.5-turbo-0613 to identify 500 samples of slang usage from LDC-AMR3.0 with the following prompt:

\textit{Please evaluate the following sentence for the presence of slang expressions.
      A slang expression is a phrase or expression that is in the online slang dictionaries
      and has a meaning that is very different from its literal form. For instance,
      `raining cats and dogs' is slang, while `middle school' is not. Although
      `middle school' is a compound phrase, it does not carry a meaning beyond
      its literal interpretation. Here is the sentence for your analysis: {premise}.
      Please format your response as follows:
      `{{Yes or No}}, {{slangs}}.'\\
      If there's no slang used, just answer `No'. If there are multiple slang expressions,
      please separate them with a semicolon (`;').
      Remember, the idioms we are interested in are those that, when taken literally,
      would have a completely different semantic meaning.}

Then we mannually check whether the extracted expressions are slang and are appropriate.
Consistent with the spirit of \cite{zhang2019paws}, we use the following prompt to query gpt-3.5-turbo-0613 to generate one paraphrase and one non paraphrase of each sentence.

{\textit{Rewrite the following sentence in two ways
Sentence: {{sentence}{}
1. Replacing ``{{slang}}'' with its intended meaning.
2. Replacing ``{{slang}}'' with its literal meaning, such that the sentence loses its original meaning.
Do not change anything else}}

Lastly, for each pair of (original\_sentence, (non)paraphrase\_sentence), we give (original\_sentence, original\_amr, (non)paraphrase\_sentence) to gpt-3.5-turbo-0613, and ask it to generate (non)paraphrase\_amr by minimally modifying the original\_amr. The prompt is as follows:

{\textit{``The AMR of the sentence `{{og\_sentence}}' is\\{{{og\_amr}}}\\ What is the AMR of the sentence `{{paraphrase}}'?\\Modified the given AMR to fit the sentence `{hypothesis}' and words not present in the sentence '{{hypothesis}}' should not appear in your AMR.\\Start you response with `('.''}}

\subsection{Intuition of Why AMR Might Be Helpful for NER}\label{appd:intuition}
For some intuition of why we choose the NER task out of the OntoNotes 5.0 dataset, it has already been shown in existing work that AMR can help event extraction \citep{Garg2015Extracting,huang2018zero}, which is also a type of named entities. Specifically, the graphical structure and typed tags of AMR make it easy to identify named entities.
For instance, 
in the sentence ``the top money funds are currently yielding well over 9\%'' in the AMR-NER dataset, we discover the AMR substructure ``(p / percentage-entity~10 :value 9),'' which makes it easy to identify ``9\%'' as a named entity of type \textit{percent}.

\section{Explanation of Linguistic Features}
\label{sec:feature_explanation}

In our analysis, we delve into specific linguistic features that exhibit strong correlations with AMR helpfulness, as detailed in \cref{tab:positive_corr_feature,tab:negative_corr_feature}.

\paragraph{Number of Adjuncts} This feature involves counting words that serve as modifiers to nouns, pronouns, verbs, and other parts of speech. Adjuncts typically provide additional context or emphasis but can be omitted without altering the core meaning of the sentence. For example, in “John really likes apples,” the word “really” is an adjunct, modifying the verb “likes.”

\paragraph{Word Complexity} We assess word complexity using the \textit{age of acquisition} metric, following the methodology of \citet{Kuperman2012Ageofacquisition}.

\paragraph{Number of Quantifier Phrase Modifiers} This feature quantifies the modifiers within quantifier phrases that adjust the head, or primary element, of the phrase. An illustration of this can be seen in the sentence “About 5000 people attended the conference,” where “about” modifies the quantifier “5000.” This concept is further explained by \citet{de-marneffe2008stanford}.

\section{Additional Experiments}
\subsection{Statistical Significance Tests}\label{appd:stats_signi}
We conduct statistical significance tests for the experiments in the main paper comparing \basemodel and \ourmodel, including \cref{tab:main_res,tab:slang_res,tab:amr_ner}. Using the Student's t-test \cite{student1908probable}, we report the significance test results in \cref{tab:stat_sig}. 
The results echo our earlier observation that the changes by \ourmodel is mostly small-scale fluctuations, as the differences are statistically insignificant in most cases.

\begin{table}[ht]
    \centering
    \small
    \setlength\tabcolsep{2pt}
    \begin{tabular}{lccccccccc}
    \toprule
    Dataset 
    & \basemodel & $\Delta$\ourmodel & Sig. ($p$)
    \\ \midrule
    PAWS & 78.25 & -3.04 & \cmark (5.209e-8)\\
    WMT & 27.52 & -0.83 & \xmark (0.0716)\\
    Logic & 55.61 & -0.49 & \xmark  (0.7300) \\
    Pubmed45 & 39.65 & -3.87 & \cmark  (0.0309)\\
    SPIDER & 43.78 & +0.61 & \xmark (0.4362)\\
    GoldSlang-ComposedAMR &86.83&-6.63 & \cmark (0.0014)\\
    GoldAMR-ComposedSlang &77.69&-8.78 & \xmark (0.1309)\\
    AMR-NER (Gold AMR) & 60.51 & +0.03 & \xmark (0.9935)\\
    AMR-NER (Parser AMR) & 60.51 & +1.91 & \xmark (0.6227) \\
    \bottomrule
    \end{tabular}
    \caption{
    For all the experiments comparing \basemodel and \ourmodel using GPT-4 mentioned in the main text, we calculate whether the difference $\Delta$\ourmodel is statistically significant (Sig.) using t-test \cite{student1908probable} by the threshold $p=0.05$, and report the actual $p$ values.}
    \vspace{-.5em}
    \label{tab:stat_sig}
\end{table}

\subsection{Additional Evaluation Results for Machine Translation}

\begin{table}[!ht]
    \centering \small
    \setlength\tabcolsep{4pt}
    \begin{tabular}{lccccc}
    \toprule
    \multirow{2}{*}{Model}     &  \multicolumn{2}{c}{BERTScore}&  \multicolumn{2}{c}{spBLEU} \\ 
    & \basemodel & $\Delta$\ourmodel & \basemodel & $\Delta$\ourmodel \\ \midrule
text-d.-001 &90.48 & 	-1.09 & 30.29 & -4.19 \\
text-d.-002 &91.00 & 	-0.30 & 33.14 & -1.67 \\
text-d.-003 &91.37 & 	-0.25 & 34.75 & -1.55 \\
GPT-3.5 &91.70 & 	-0.06 & 37.09 & -0.43 \\
GPT-4 &91.79 & 	-0.08 & 37.71 & -0.72\\
    \bottomrule
    \end{tabular}
    \caption{Performance on WMT by additional metrics, BERTScore and spBLEU.}
    \label{tab:translation}
\end{table}

In the main paper, we mainly report the performance of machine translation using the standard evaluation metric BLEU \cite{papineni2002bleu}. Recent studies has proposed new metrics to evaluate the quality of machine translation, such as BERTScore \cite{zhang2020bertscore} and spBLEU \cite{goyal-etal-2022-flores}, so we also report the model performance according to these two additional metrics in \cref{tab:translation}. We use the version of spBLEU built from the Flores-200 dataset. \footnote{\url{https://github.com/facebookresearch/flores/tree/main/flores200}} The performance trend is consistent with \cref{sec:q1}, where AMR has a marginal effect on the baseline LLM performance.

\subsection{Larger-Scale Ablation Study Using WMT}\label{appd:ablation}
\begin{figure}[ht]
\centering
\begin{tikzpicture}
\begin{axis}[
    xlabel={
    AMR/Text Tokens (\%) Kept in the Prompt},
    ylabel={Task Performance (BLEU)},
    xmin=0, xmax=100,
    ymin=0, ymax=40,
    ymajorgrids=true,
    grid style=dashed,
    width=0.6\columnwidth,
    tick label style={font=\footnotesize},
    legend style={at={(1.02,0.2)}, anchor=west, font=\footnotesize},
    legend style={font=\footnotesize, cells={align=left}},
    label style={font=\footnotesize}
]

\addplot[
    color=orange,
    mark=*,
]
coordinates {

    (0,4.72)
    (20,1.94)
    (40,4.55)
    (60,10.94)
    (80,19.95)
    (100,31.91)

};
\addlegendentry{Text}
\addplot[
    color=blue,
    mark=square,
]
coordinates {

    (0,32.81)
    (20, 32.37)
    (40,32.20)
    (60,32.43)
    (80,31.90)
    (100,31.91)

}; %
\addlegendentry{AMR}

\end{axis}
\end{tikzpicture}
    \caption{Ablation studies of AMR and text representations in the prompt on 1,000 random samples of the WMT dataset using {GPT-4}.
    Starting from the \ourmodel prompt with the complete text and AMR, we randomly drop out a certain portion of tokens in the text/AMR, and see the effect on the task performance. 
    }
    \label{fig:amr_text_ablation_newstest}
\end{figure}
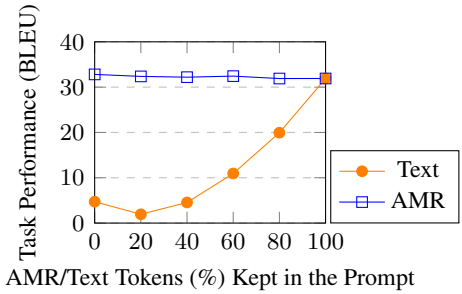
We understand that the ablation study in \cref{sec:ablation} is on a small scale (in order to use the gold annotated data). As an alternative tradeoff to regress a bit on the data quality, but aim at a larger scale, we also conduct the same ablation study on 1,000 random samples from the WMT dataset using predicted AMRs. Our results in \cref{fig:amr_text_ablation_newstest} also confirm that text has a more instrumental role for LLMs.

\subsection{Few-Shot Experiments}
\label{appd:fewshot}
In addition to the main results of the overall effect of AMR in \cref{sec:q1} using zero-shot prompting, we also check whether adding few-shot examples to the prompt will help. Since the experiments are very costly, we conduct a small-scale preliminary check running few-shot \ourmodel on 200 random samples from PAWS using \texttt{gpt-3.5-turbo-0613}. As AMRs are lengthy, and PAWS is a binary classification task, we select one random example with the positive label, and another with the negative label, totaling an average prompt length of 371 tokens. The resulting F1 score is 63.67, close to \basemodel performance of 63.92, serving as a preliminary observation that the few-shot setting might not change our observations much.

\begin{table}[ht]
\label{tab:eg_text_amr}
    \centering \tiny
    \setlength\tabcolsep{2pt}
    \begin{tabular}{lp{6cm}}
\toprule

\multicolumn{2}{l}{\textbf{\textit{Paraphrase Detection (PAWS)}}} \\
Text & Sentence 1: The defendant then broke into the house and tried unsuccessfully to remove and open the vault. \newline
Sentence 2: The defendant then broke into the house and tried unsuccessfully to open the safe and then to remove them.
\\
AMR & AMR1: (a / and~7 :op1 (b / break-02~3 :ARG0 (d / defendant~1) :ARG1 (h / house~6)) :op2 (t2 / try-01~8 :ARG0 d :ARG1 (a2 / and~12 :op1 (r / remove-01~11 :ARG0 d :ARG1 (v / vault~15)) :op2 (o / open-01~13 :ARG0 d :ARG1 v)) :ARG1-of (s / succeed-01~9 :ARG0 d :polarity -~9)) :time (t / then~2))  \newline
AMR2: (a / and~7 :op1 (b / break-02~3 :ARG0 (d / defendant~1) :ARG1 (h / house~6)) :op2 (t2 / try-01~8 :ARG0 d :ARG0-of (s2 / succeed-01~9 :polarity -~11) :ARG1 (a2 / and~14 :op1 (o / open-01~11 :ARG0 d :ARG1 (s / safe~13)) :op2 (r / remove-01~17 :ARG0 d :ARG1 s))) :time (t / then~2)) 
\\
\midrule
\multicolumn{2}{l}{\textbf{\textit{Translation (WMT16)}}}\\
Text & Obama receives Netanyahu
\\
AMR & (r / receive-01~1
    :ARG0 (p / person~0
        :name (n / name~0
            :op1 "Obama"~0))
    :ARG1 (p2 / person~2
        :name (n2 / name~2))
    :rel (e / Netanyahu~2))
\\
\midrule
\multicolumn{2}{l}{\textbf{\textit{Logical Fallacy Detection (Logic)}}}\\
Text & On my walk to work this morning, a woman on her bike nearly ran me off the sidewalk. I hadn't realized that cyclists were so aggressive and rude!
\\
AMR & (m2 / multi-sentence~19 :snt1 (n / near-02~13 :ARG2 (r3 / run-10~14 :ARG0 (b / bike~12 :poss (w2 / woman~9)) :ARG1 (i / i~15) :ARG2 (s / sidewalk~18) :time (w / walk-01~2 :ARG0 i :ARG2 (w3 / work-01~4 :ARG0 i) :time (d / date-entity~6 :dayperiod (m / morning~6) :mod (t / today~5))))) :snt2 (r / realize-01~23 :ARG0 (i2 / i~20) :ARG1 (h / have-degree-91~27 :ARG1 (p / person~25 :ARG0-of (c / cycle-01~25)) :ARG2 (a2 / and~29 :op1 (a / aggressive~28) :op2 (r2 / rude-01~30 :ARG1 p)) :ARG3 (s2 / so~27)) :polarity -~22)) 
\\
\midrule
\multicolumn{2}{l}{\textbf{\textit{Event Extraction (Pubmed45)}}}\\
Text & Examination of activated phospho-MEK levels revealed that the FBm and S729 mutations had no effect on MEK activation induced by the high-activity V600E B-Raf protein; however, the FBm and S729A mutations increased and decreased, respectively, the abilities of the intermediate G466A and kinase-impaired D594G B-Raf proteins to activate MEK (Fig. 4B), indicating a correlation between the transformation potential of these proteins and their ability to activate ERK cascade signaling in vivo.
\\
AMR & (c3 / contrast-01~35 :ARG1 (r / reveal-01~7 :ARG0 (e4 / examine-01~0 :ARG1 (l / level~6 :ARG1-of (a / activate-01~2) :mod (e / enzyme~5 :name (n / name~5 :op1 "MEK"~5) :ARG1-of (p / phosphorylate-01~3)))) :ARG1 (a6 / affect-01~17 :ARG0 (m / mutate-01~14) :ARG1 (a2 / activate-01~20 :ARG1 (p3 / protein~19 :name (n5 / name~19 :op1 "MEK"~19)) :ARG1-of (i4 / induce-01~21 :ARG0 (p4 / protein~33 :name (n2 / name~5 :op1 "FBm"~10 :op1 "V"~27 :op2 600~28 :op3 "B-Raf"~30) :ARG0-of (a5 / activity-06~26 :ARG1-of (h / high-02~24))))) :polarity -~16) :ARG1-of (d2 / describe-01~76 :ARG0 (f / figure~73 :ARG0-of (i3 / indicate-01~78 :ARG1 (c4 / correlate-01~80 :ARG1 (p2 / potential~84 :domain (t2 / transform-01~83 :ARG0 (p7 / protein~87 :mod (t / this~86)))) :ARG2 (c2 / capable-01~90 :ARG1 p7 :ARG2 (a4 / activate-01~92 :ARG0 p7 :ARG1 (s / signal-07~95 :ARG0 (e3 / enzyme~93 :name (n7 / name~93 :op1 "ERK"~93 :op2 "cascade"~94))) :manner (v / vivo~97))))) :mod 4B~75))) :ARG2 (a7 / and~45 :op1 (i2 / increase-01~44 :ARG0 (m2 / mutate-01~43 :ARG1 (e2 / enzyme~19 :name n2)) :ARG1 (c / capable-01~51 :ARG1 (a8 / and~58 :op1 m2 :op1 (p5 / protein~68 :name n2) :op2 (m3 / mutate-01~43 :ARG1 p3) :op2 (p6 / protein~68 :name (n6 / name~68 :op1 "G"~55 :op2 466~56 :op3 "A"~57) :ARG2-of (i / impair-01~61 :ARG1 (k / kinase~59)))) :ARG2 (a3 / activate-01~70 :ARG0 a8 :ARG1 e2))) :op2 (d / decrease-01~46 :ARG0 m3 :ARG1 c)) :rel (n3 / name~5 :op1 "D"~62 :op2 "594"~63) :rel (n4 / name~12 :op1 "S"~12 :op2 "729"~13) :rel (f2 / figure~73) :rel (i5 / intermediate~54)) 
\\
\midrule
\multicolumn{2}{l}{\textbf{\textit{Text2SQL (SPIDER)}}}\\
Text & List the number of all matches who played in years of 2013 or 2016.
\\
AMR & (l / list-01~0     :ARG0 (y / you~0)     :ARG1 (n / number~2         :quant-of (m / match-03~5             :ARG0-of (p / play-01~7                 :time (o / or~12                     :op1 (d / date-entity~11                         :year 2013~11)                     :op2 (d2 / date-entity~13                         :year 2016~13)))             :mod (a / all~4)))     :mode imperative~0)
\\
\midrule
\multicolumn{2}{l}{\textbf{\textit{Named Entity Recognition (AMR-NER)}}}\\
Text & Then, in the guests' honor, the speedway hauled out four drivers, crews and even the official Indianapolis 500 announcer for a 10-lap exhibition race.
\\
AMR & (h / haul-01~10 :purpose (h2 / honor-01~6 :ARG1 (g / guest~4)) :purpose (r / race-02~29 :ARG1-of (e3 / exhibit-01~28) :ARG3 (l / lap~27 :quant 10~25)) :ARG0 (s / speedway~9) :ARG1 (a / and~14 :op1 (p / person~13 :quant 4~12 :ARG0-of (d / drive-01~13)) :op2 (c / crew~15) :op3 (p2 / person~15 :ARG0-of (a2 / announce-01~22 :ARG1 (e2 / event~21 :name (n / name~20 :op1 "Indianapolis"~20 :op2 500~21))) :mod (e / even~17) :mod (o / official~19))) :direction (o2 / out~11) :time (t / then~0)) 
\\
\bottomrule
    \end{tabular}
    \caption{Example text-AMR pairs for each task.}
\end{table}
\end{document}